# A Process Planning System with Feature Based Neural Network Search Strategy for Aluminum Extrusion Die Manufacturing

S. Butdee[1], C. Noomtong[1], S. Tichkiewitch[2]
[1] IMSRC, Department of Production Engineering, Faculty of Engineering, KMUTNB, Bangkok, Thailand
[2] G-SCOP Laboratory, Grenoble Institute of Technology, Grenoble, France

*Abstract*

*Aluminum extrusion die manufacturing is a critical task for productive improvement and increasing potential of competition in aluminum extrusion industry. It causes to meet the efficiency not only consistent quality but also time and production cost reduction. Die manufacturing consists first of die design and process planning in order to make a die for extruding the customer's requirement products. The efficiency of die design and process planning are based on the knowledge and experience of die design and die manufacturer experts. This knowledge has been formulated into a computer system called the knowledge-based system. It can be reused to support a new die design and process planning. Such knowledge can be extracted directly from die geometry which is composed of die features. These features are stored in die feature library to be prepared for producing a new die manufacturing. Die geometry is defined according to the characteristics of the profile so we can reuse die features from the previous similar profile design cases. This paper presents the CaseXpert Process Planning System for die manufacturing based on feature based neural network technique. Die manufacturing cases in the case library would be retrieved with searching and learning method by neural network for reusing or revising it to build a die design and process planning when a new case is similar with the previous die manufacturing cases. The results of the system are dies design and machining process. The system has been successfully tested, it has been proved that the system can reduce planning time and respond high consistent plans.*

## 1 INTRODUCTION

Aluminum extrusion is a hot deformation process used to produce long, straight, semi finished metal products such as bars, solid and hollow sections, tubes and many shapes products. The hot aluminum extrusion process is pressed under high pressure and temperature in a specific machine. A billet is squeezed from close container to be pushed through a die to reduce its section [1]. A profile shape is deformed in order to be adapted to the die orifice shape. The significant machine used to press aluminum is called an aluminum extrusion press. The main tooling of aluminum extrusion process is a die. It is used to form aluminum profile shapes. Hence, die must be efficiently constructed to support aluminum extrusion process to ensure obtaining a good extruded profile. Aluminum extrusion die manufacturing is the work significant activity in aluminum extrusion industry in order to obtain an efficiency die. Die manufacturing is proceed of two main phases, including the die design and the process planning for machining to make a die. Die design is based on the skill of a die designer who has accumulated the working experiences for many years. Die designer gives the die design concept and details to create the die geometry in CAD system. In addition, process planning for die machining is also based on the experience of a die process planner expert. The knowledge base of die design and process planning should be captured from the experts. It would be formulated, managed and stored in a computer system in order to use it and revise with new situation die design. At the present, intelligence system has been discussed in knowledge management for industrial fields, such as design and manufacturing. Intelligent systems can be knowledge based system, genetic algorithm, rule-based and frame-based expert system, fuzzy logic, artificial neural networks which are general tools using in order to solve the complex or specific problem in engineering works [2]. For aluminum extrusion die design and process planning of die manufacturing, we propose an artificial intelligent system to aid management of the knowledge of design and process planning with the feature based neural networks technique to search the similar previous die design case, in order to reuse the previous cases in the new case design and manufacturing. Die geometry can be constructed by the combination of geometrical features. The principle features are holes, edges, grooves, pockets etc. These features are given by die designer experts. Therefore die features should be organized in feature library for reusing at a new die design and to decrease die design lead time. The knowledge of die design is translated from the human skill to the knowledge based system via feature definition. Hence, the knowledge base of die design can be directly based on from die geometry design. The geometrical features data of a die contains feature shape, dimensions and so on. Moreover, each feature of die geometry is employed to define the possibility of machining processes to fabric a die. The knowledge based system of designing of extrusion die is organized with frame-based and rule-based system. This paper consists of five sections. Firstly, is to explain the fundamental of aluminum extrusion process and extrusion tools. The second section describes the main tool used for extrusion process (aluminum extrusion die). The third section presents the principle methods of die design process, the knowledge base of die design, and die geometry. The fourth section addresses the process planning of die manufacturing, including the knowledge base of die process planning. The fifth section presents feature based neural network for aluminum extrusion die manufacturing. Case study is illustrated in the sixth section. Finally, the summary section concludes the application of the proposed system to support aluminum extrusion die manufacturing.

## 2  ALUMINUM EXTRUSION
### 2.1 Extrusion process
An extrusion press is used to extrude many materials such as lead, aluminum, copper, zinc, brass, etc.

There are four characteristic differences among the various methods of extrusion and press used:
- The movement of the extrusion relative to the stem - direct and indirect process.
- The position of the press axis – horizontal or vertical press.
- Type of drive – hydraulic (water or oil) or mechanical press.
- Method of load application – conventional or hydrostatic extrusion.

Normally, the extrusion process can be classified into two basic methods, direct and indirect extrusion. This paper we only discuss on the direct extrusion process. There is the simplest production of aluminum extrusion process, which can be carried out without lubricant. The most important and principle method used in extrusion is the direct process as shown in Figure 1, which follows the step sequence as given below [3]:
- Loading the billet and piston into the press
- Extrusion of the billet, press billet through die
- Decompression of the press and opening of the container to expose the discard and the piston
- Shearing the discard or backend or input a new billet
- Returning the shear, container and ram to the loading position

### 2.2 Equipment and tooling in aluminum extrusion process
The principle equipment and tooling in hot aluminum extrusion process are explained in [4], including:
#### 2.2.1 Press machine
An aluminum extrusion press is the main machine to extrude aluminum ingot (Billet) through a die; pressure from machine is generated by hydraulic system. Almost horizontal and hydraulic extrusion press is widely using in an aluminum extrusion industrial. Hydraulic extrusion press is illustrated in Figure 2.
#### 2.2.2 Run out table and puller
The extrusion emerges from the press onto the run out table, which supports the extrusion. A puller or pullers guide the extrusion and keep it under constant tension. Run out table and puller equipments are shown in Figure 3.
#### 2.2.3 Die oven
Die oven is an equipment to preheat die temperature on appropriate state for extrusion. In hot forming material tooling must be heated to increase mechanical properties and to control temperature environment. Normally, an aluminum extrusion die is heated nearest billet temperature about 425-450°C. Another, the significant parameter is die preheat time. To prevent under heating or over heating of die, die preheat time also needs to be strictly controlled. The recommended die preheating practices are as follow:

- Minimum soaking time: 1hr/inch of the die and backer.

- Maximum allowance time after a die reaches specified temperature
  300°C - 24 hours
  370°C - 10 hours
  420°C - 8 hours
  480°C - 2 hours

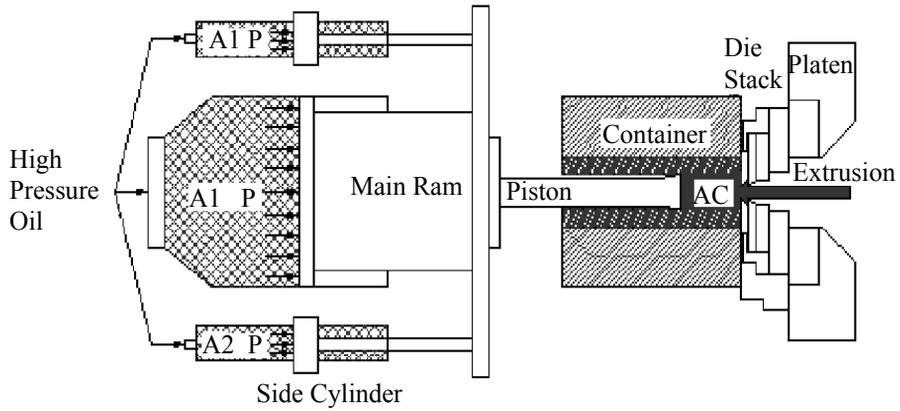

Figure 1: Direct extrusion process

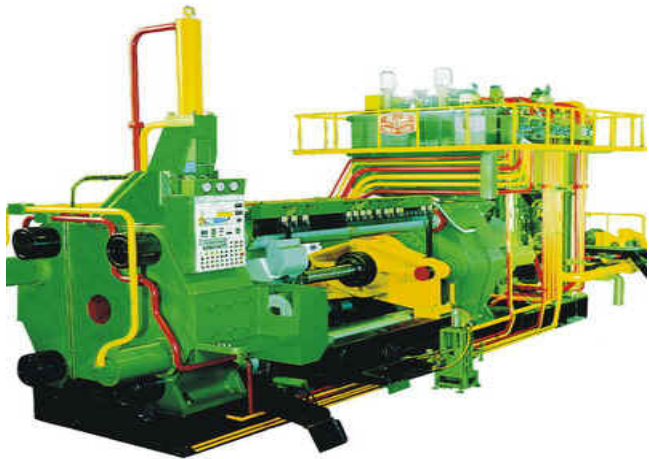

Figure 2: Press machine and schematic of direct extrusion press

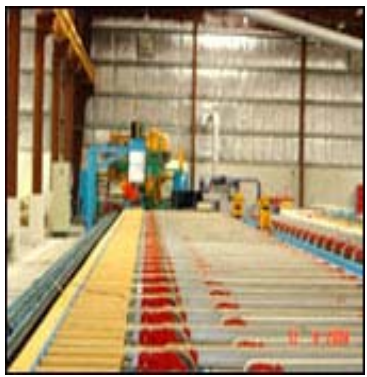 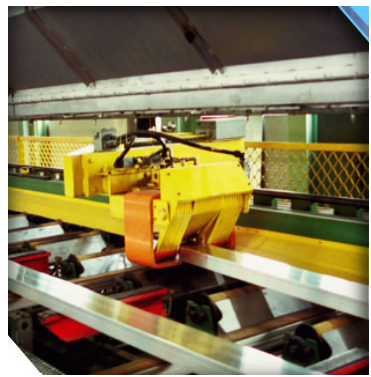

Figure 3: Run out table and puller

Indeed, die preheat time based on the efficiency of die oven and any parameter, each company should has its standard time. Die oven is illustrated in Figure 4.

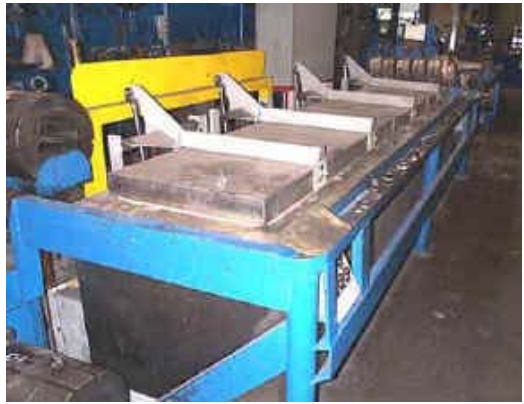

Figure 4: Die oven

### 2.2.4 Billet Oven
Billet oven is used to preheat billet temperature 420-450 °C. Most common are gas or induction ovens. Billet oven is shown in Figure 5. High billet temperatures (Temperature more than 480 °C) will reduce the extrusion pressure, but decrease extrusion speed of the extrusion in order to avoid surface defects. Globally, the productivity of the press is reduced.

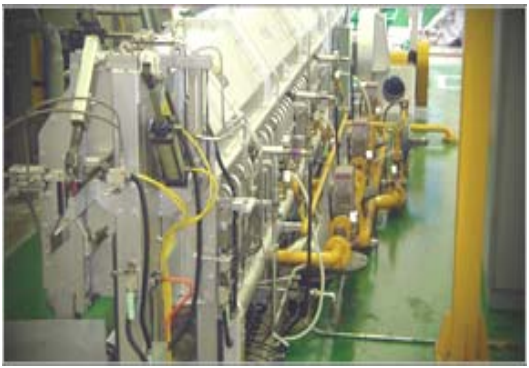

Figure 5: Billet oven

### 2.2.5 Stretcher
Stretcher is an equipment to extend aluminum profiles longitudinal direction. The purpose is to relieve residual stress in extruded profiles. Normally, stretch straight extrusion length about 0.5% and decreases the cross-sectional dimensions correspondingly, over stretching can result in distortion or dimensions out of tolerance. Stretching of 2% or more can lead to orange peel defects on the extrude surface. A high level of stretching is required for straightening of distorted sections. A better approach is to control metal flow through the die to reduce distortion rather than to apply high levels of stretching. Aluminum extruded stretcher is shown in Figure 6.

### 2.2.6 Extrusion cut-off saws
A cut-off saw is an equipment to cut the aluminum profiles to preset length. Circular blades for cutting aluminum extrusions are 16 to 20 inches and are made from carbide. Modern saws have a self-contained hydraulic system and chip collector to trap continuously. Figure 7 illustrates extrusion cut-off saws

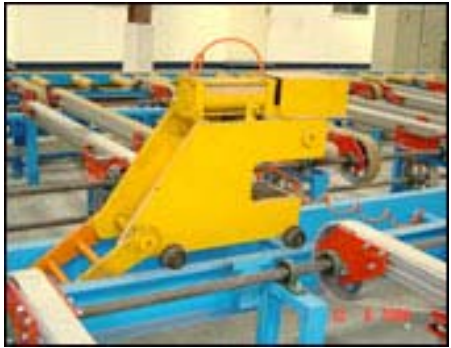

Figure 6: Stretcher

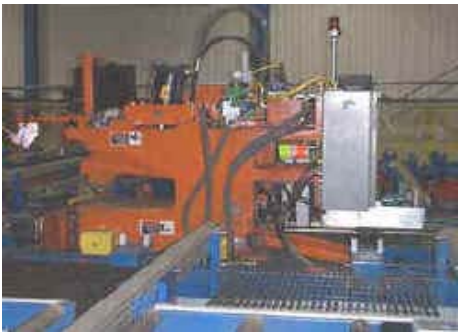

Figure 7: Extrusion cut-off saws

### 2.2.7 Aging Oven

The aging oven is given a heat treatment the extruded profiles to improve the mechanical qualities of the product. The extruded profiles are heated to a set temperature for an extended period ranging from 4-10 hours depending on mechanical properties required. Not all extrusions are artificially aged at the plant. Aging oven is demonstrated in Figure 8.

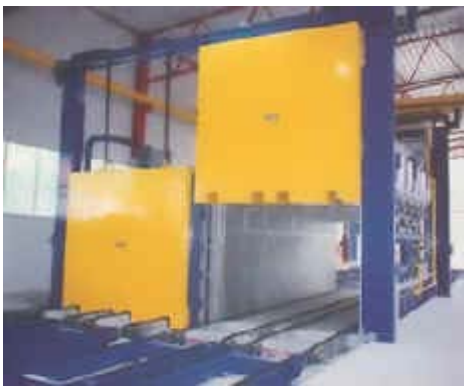

Figure 8: Aging oven

## 3  ALUMINUM EXTRUSION DIE

In aluminum extrusion process, a die is an important tool to deform aluminum ingot (Billet) to get straight aluminum profiles. An aluminum profile is formed by die orifice (die hole). In addition, the quality of products and extrusion productivity often depend on die performance. An aluminum extrusion die can be classified into three basic types: solid, semi-hollow and hollow die. Its can produce aluminum profiles solid, semi-hollow and hollow section respectively. The normal type of die for solid and open semi-hollow shapes can be used for all metals that are extruded. Hollow sections in a variety of shapes and sizes are reserved for aluminum alloys apart from the simple hollow sections that can be produced in heavy metals or steel with a mandrel. The development of die design for simple and complicated aluminum sections is determined first of all because the extruded products have significant position in the world market. Aluminum extrusion tooling includes die set, die holder, bolster/sub-bolster or pressing ring, etc. The typical aluminum extrusion dies are illustrated in Figure 9. Die set of solid and hollow die are presented as in Figure 10 and 11 respectively. In addition, the table 1 explains the fundamental function of each die part component to play an important role in aluminum extrusion process.

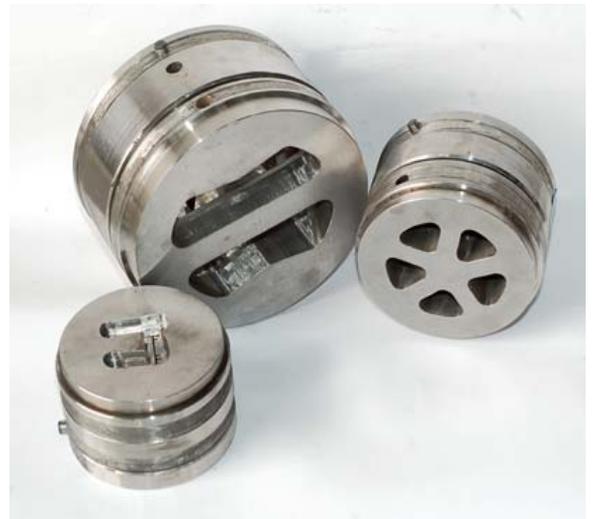

Figure 9: Aluminum extrusion dies

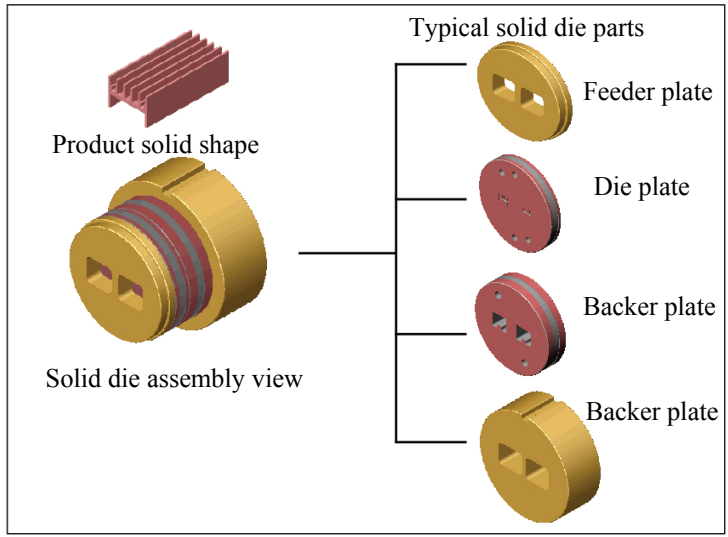

Figure 10: Solid die components

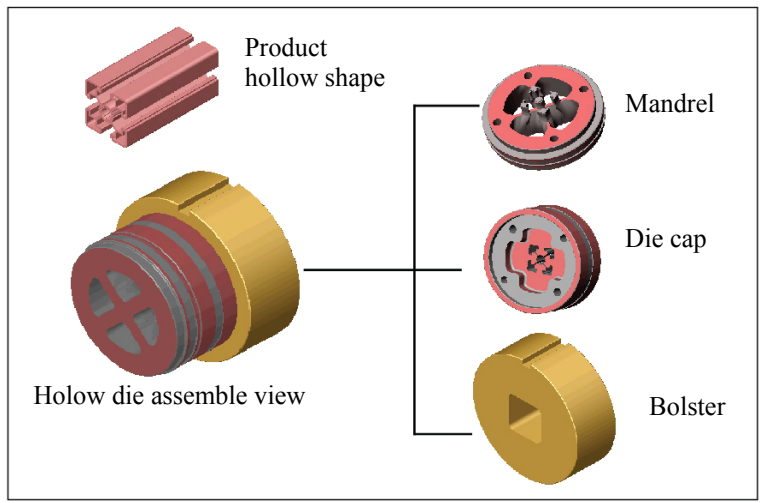

Figure 11: Hollow die components

Table 1: The functions of extrusion tooling

| Die | Form the section. A solid die type is called die plate and for hollow die, we found its die cap and mandrel. |
|---|---|
| Backer | Supports the tongue of die to prevent collapse or distortion. The shape of backer orifice is often closely related to die orifice. |
| Feeder plate | Balance aluminum flow through die orifice. |
| Bolster | Supports extrusion load is transmitted from die and backer. |
| Die holder | Holds the die and, to some degree, the die backer. |
| Die carrier | Holds the die set in the press. |
| Bridge | Divided metal flows and supports the mandrel. |

The difficulty to obtain a good die is on the equal section. It is the condition to obtain a good straight bar (see 4.1). A die for an aluminum extrusion needs properties follow as:

- Accurate dimensions and product shape, to avoid the need for any corrective work and to decrease cost repair die.
- Maximum possible die life, increase productivity and decrease extrusion cost per unit.
- Maximum length of the extruded section, unless it is determined by calculating extrusion yields and selecting a proper billet sizes.
- High extrusion speed, depend on die design and extruded operation.
- A good-quality surface finish product.
- Low manufacturing cost.

These requirements are usually fulfilled with rod and simple shapes. However, as the die more complex, it becomes increasingly more difficulty to comply six requirements. Many factors have to be considered in the design and construction of a die, including the flow pattern, maximum specific pressure, geometrical shape of the section, wall thickness and tongue sizes, shape of bearing surface, and tolerance of the section. Extrusion dies are essentially thick, circular steel disks containing one or more orifices of the desired profile. A die type and die geometry depend on the characteristics of each profile to be extruded. There are normally fabricated from hot working steel, as H-13 (American Standard) and heated to the desired condition. Such difficult section of a bar served as example of what it formed to do for forging exhibition. In a typical extrusion, the extrusion die will be placed in the extrusion press along with several supporting tools. These tools, also repetition of the velocity in each point of the final made from hardened tool steel, are known as backers, bolsters and sub-bolster. Its can provide support for the die during the extrusion process to prevent die broken and to extend the die life. In addition die support tools contribute to improved tolerance controls and extrusion speed. A tool stack for a hollow die is similar to that used for a solid die. A hollow die is a two-piece construction, one piece forming the inside of the hollow profile and the other piece forming the outside of the profile. It likewise requires the use of additional support tools.

### 3.1 Solid dies

Solid dies are used to produce profiles that do not contain any voids. Various styles of solid dies are used, depending on the equipment and manufacturing philosophy of the extruder. Some prefer to use recessed pocket or weld-plate style dies. A pocket die has a cavity slightly larger than the profile itself, approximately to deep. This cavity helps control the metal flow and allows the billets to be welded together to facilitate the use of a puller. Both pocket and weld plate type dies provide for additional metal flow control, compared to that of the flat-face type die. A weld plate is a steel disk that is placed (often pinned and/or bolted) in front of a solid die. It has an opening that controls the flow of metal to the orifice. Weld or feeder plates serve to control contour, and/or spread the aluminum. It is also important that the layout of a multi-hole die is arranged to prevent extrusions from rubbing together or running on top of each other as they leave the press. A flat surface and not the edge of a leg or rib should run along the run-out table to prevent small part of profiles to be bent. In Figure 12 illustrates typical solid dies one and four profiles for a die plate.

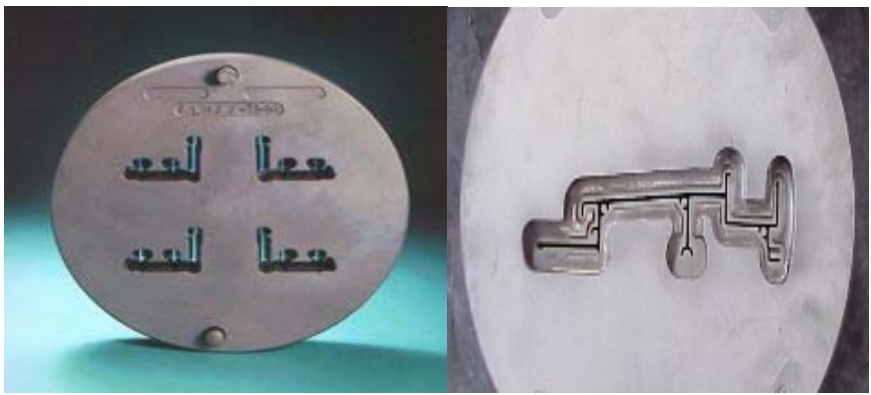

Figure 12: Typical solid dies

## 3.2 Hollow dies
A hollow die produces profiles with one or more voids such as tube products. The profiles could be as simple as a tube with one void or as complex as a profile with many detailed voids. The most common type of hollow die is the porthole die, which consists of a mandrel and cap section; it may or may not have a backer or back plate. The mandrel, also known as the core, generates the internal features of the profile. The mandrel has two or more ports, it based on the shape of profiles or die design method. The aluminum billet will be separated into each port and rejoins in the weld chamber prior to entering the bearing area and die orifice. Webs, also known as legs, which support the core or mandrel section, separate the ports. The cap creates the external features of the profiles. It is assembled with the mandrel. The die cap has a pocket or welding chamber, which is rejoin metal flow from porthole through die orifice. It is assembled with the mandrel. The backer or back plate, when used, provides critical tool support and it is immediately adjacent to, and in direct contact with the exit side of the cap. In case the profiles have critical point a backer will necessary be supported to avoid die broken and to extend die life. Typical hollow dies are presented in Figure 13.

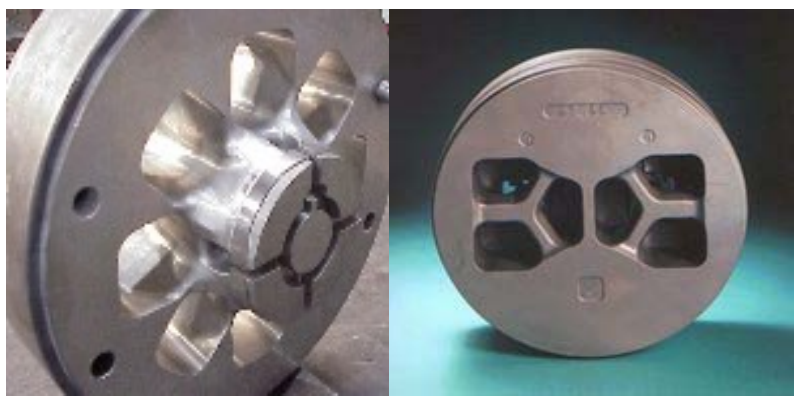

Figure 13: Typical hollow dies

## 3.3 Semi-hollow dies
A semi-hollow die is used to produce profiles having semi-hollow characteristics as defined in Aluminum Standard and Data (published by the Aluminum Association, Inc.). Semi-hollow dies have port holes, legs, bridges as same as hollow die but without cores to make a void of section. It gives the extruded solid profile shape with high tongue ratio. In practice, semi-hollow profile, tongue ratio is grater than 5.

The semi-hollow classification derives from a mathematical comparison between the area of the partially enclosed void and the mathematical square of the size of the gap. This ratio (area/gap²) is called the tongue ratio. Depending on the tongue ratio, semi-hollow dies can be constructed as flat, recessed-pocket, weld-plate, or porthole design. Porthole dies are more prevalent in the production of semi-hollow profiles.

## 4 ALUMINUM EXTRUSION DIE DESIGN
### 4.1 Die design process
Die design is one of crucial task in aluminum extrusion process. The traditional die design is based on skills and experiences of a die designer. He/she learns and accumulates knowledge of die design from the previous die design cases. The success and failure cases are studied and analyzed in order to use their information from these cases for improving a new die design to avoid failure case. In practice, a die may be tested many times before the extrusion profile is satisfactory [5]. The first die test points out the efficiency of die design. If the extruded profile shape is not perfect, the tester will modify die geometry such as bearing length to balance extrusion speed of the section for obtaining profile shape perfectly.

The quality factors of die design are: decrease the number of tests, increase die life, produce a good product and provide high productivity. This part describes the fundamental of aluminum extrusion die design process. In general, die design process calculating shrinkage all point on profile section, determining feed or porthole shapes (will be defined just after) and sizes for controlling aluminum flow, determining dimensional die orifice and tongue deflection, calculating the bearing lengths, designing tooling support. Furthermore, a die designer has to consider the size of the containers of the machines and the number of extruded profiles graved in the same die for each size of container. Each die design process should be focused on the parameters that have effect to die efficiency such as shape factor, extrude ability, extrusion ratio, etc. In fact, die design or die geometry will be modified following the profile characteristics, especially, profile shape and dimensions. To obtain high extrusion productivity, die design must be optimized to increase the capability of the extrusion process and to overcome die fail. In addition, a tester should have a through understanding of the different functions of each die consists of laying out, selecting type of die and size, features (i.e. leg, sink in, porthole, bearing etc.). These features will be modified to interact with the process parameters and each of the product characteristics. The complex flow of material in the extrusion die or container often creates different deformation conditions in various regions of an extruded product. Die design has to imagine the metal flow pattern in die and container in order to make decisions for the definition of each die feature. Especially, die orifice and bearing length are the important features, which have directly effect with metal flow velocity. In theory, flow velocity of each point on the section should be balanced to avoid the extruded profile twist. Figure 14 illustrates the typical profile, which is incompletely deformed shape because die bearing lengths are unbalance then the velocities of two sides of the section are large different.

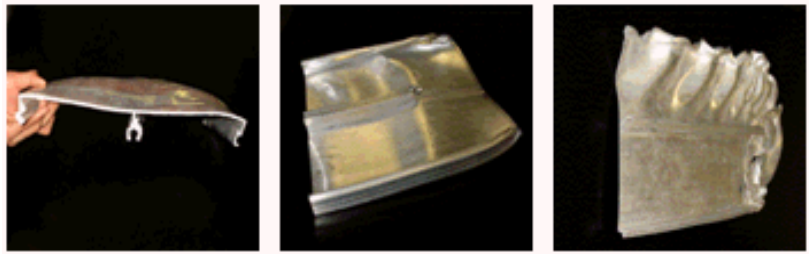

Figure 14: Typical profiles unbalance shape

As already described, bearing length is used to control the extruded profile shape and extrusion speed. The suitable given bearing length is required in aluminum extrusion die design. If bearing length is too short, the profile shape may be twist or wave. On the other hand, great bearing length will reduce extrusion speed and productivity. In addition, the extruded surface quality is not satisfactory.

### 4.2 The knowledge of die design

In general, die design is based on skills and experiences from a die designer, who accumulates the knowledge of die design for using this knowledge to solve the die design problems. The problems in die design work are: the selection of the suitable press, the type of die, the choice of material, the laying out the profile(s) on the die in order to optimize the extrusion yield, the different of the contraction cooling of each sections, the unbalance velocity of the metal flow on each point of the sections, the consolidation of the tooling to withstand in extrusion process, etc. These problems have to use the knowledge base of die design in order to find the best solution for solving each problem and to approach a good die.

Figure 15 shows the basic flat die for solid sections, consisting of feeder plate, die plate, backer plate, bolster, and sub-bolster. Naturally, the metal flows from the container into a die by passing feeder, die orifice, back opening, bolster, and sub-bolster respectively. The feeder plate contains one or more cavity to be used to weld the metal flow together before through into die orifice and to balance the metal flow to the die. Sometime, flat die design is without feeder plate; if we can sink the face of the die around the cavity in the die plate.

However, the feeder plate can permit to enable successive billets to weld together and to enable wider sections to be extruded than is normally possible from any given container size. The die plate is used to deform the material in order to take the shape of die orifice (see Figure 15). The backer die deflection and to minimize the die stresses. In case of no long tongues, the simply sections are usually extruded by using one of a standard range of backers. If the shape contains critical sections that provides the immediate support to the die to reduce need good support, then a custom backer with an aperture closet the die orifice can be used for supporting the tongue necessary.

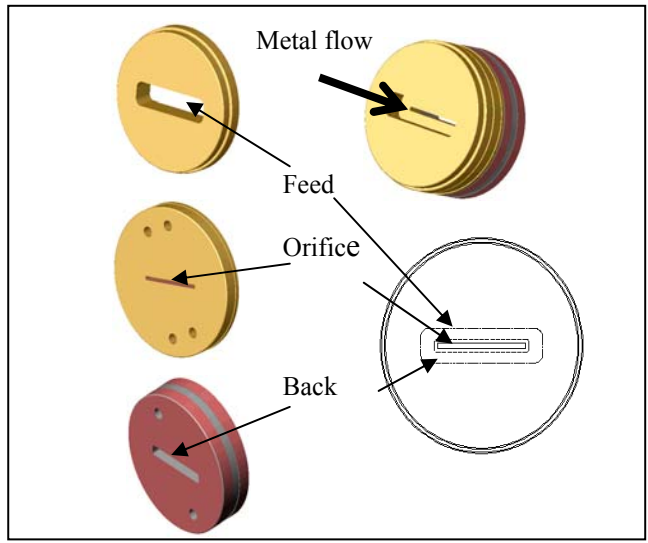

Figure 15: Solid die geometry and features

Furthermore, bolster and sub-bolster are used to support consolidation of a die. The bolster supports the backer and usually sufficient support can be achieved by using standard bolster.

Porthole dies are primarily used to produce hollow extrusions as shown in Figure 16. The basic tooling in hollow extrusions composes of mandrel, die cap, bolster, and sub-bolster. The porthole dies are suitable for multi-holes dies and can also be used with the maximum section circumscribing circle diameter relative to the container diameter. A favourable extrusion ratio can be selected by using the optimum material flow; heating to relatively low temperatures should then be sufficient. Moreover, this type of die is used for sections with very critical tongues and cross sections. However, the die is manufactured from a single piece of steel, die correction and die cleaning by polishing are difficult. Mandrel is the projection, fixed or floating that is positioned in front of the die cap in order to make the empty part of the sections. In porthole die the aluminum flow is split by legs, which support the core of the mandrel. The material flows around the legs, through the feeder holes and is welded together in the welding chamber as shown in Figure 16. The profile shape is formed by the clearance between the core on the mandrel and the die opening in the die cap. The wall thickness of the extrusion is determined by the difference in the diameters of the die apertures in die cap and the mandrel. Die cap contains weld chamber, die orifice, bearing length. The hollow shapes are formed by the core and bearing surface on die cap. In practice, the distance between core and die orifice is given by determining the contraction cooling of the sections. Bolster and sub-bolster for hollow dies have the functionality than for solid dies, in order to protect the die collapse.

Besides, the fundamental of die design, a die designer needs the explicit and implicit knowledge during die design stages in order to obtain the efficiency of the die design. This knowledge encompasses the selection the suitable press and tooling, the determination number of die opening and the laying out of the die orifices, the calculation shrinkage allowances on section dimensions, the providing of additional dimensional allowances for die dishing and tongue deflection, the determination of the bearing length to control the profile speed in each point on the sections, and the checking of the adequate support for critical sections. The knowledge about die design is described the next part.

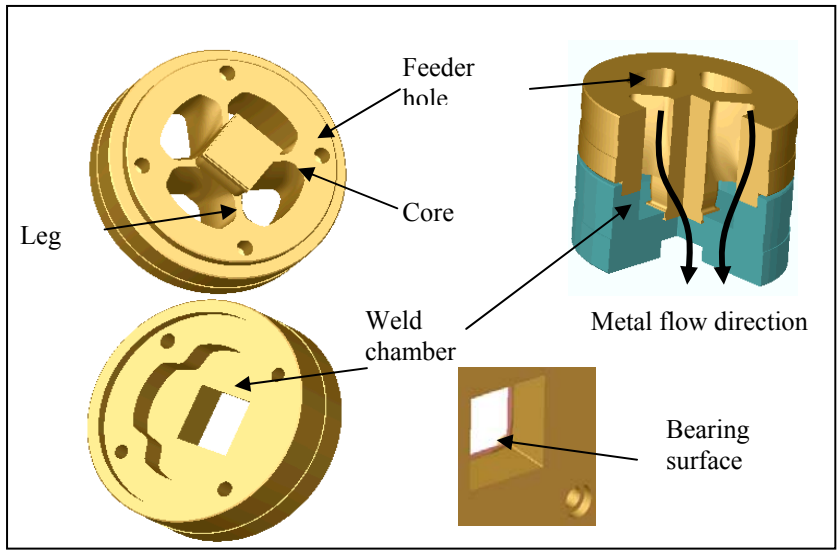

Figure 16: Hollow die geometry and features

### 4.3 Die geometry

Die can be distinguished into three categories as solid, hollow and semi-hollow die as described above. The difference die geometries provide the various shapes of the profiles. Each die part has different machining processes in order to perform the final die shape. We proposed that the features of die parts are constructed based on OO (Objected-Oriented). Classes describe the feature objects and it can be inherited by deriving a feature subclass from feature head class. Each class has methods and attributes to describe their class entity. These feature classes are managed in database system. By example we have a holes feature class and then it can be derived to blind hole, through hole, counter bore, countersink, etc. A user defines these features data and translates geometry data from CAD into feature classes for using in process planning. For instance, class hole has methods to calculate holes surface and volume, and attributes of holes feature includes diameter, depth, centre point, axis, tolerance, and surface finishing. Each attribute is used to define holes entity and to calculate holes surface and volume respectively. Moreover, it is used to select machining tools, such drill, centre drill, tap, and so on to make the holes. Figure 17 illustrates the typical features of die geometry for principle solid die, for the It has three parts: feeder, die, and back plate.

A feature is a basic entity or information that has properties to involve in engineering design and manufacturing. It also represents engineering significance, not only geometric and topologic information. Non-geometric property such as, roughness, hardness of material, and so on are presented [6]. Die geometry features are used to support die design in order to generate die geometry. It can decrease die design lead-time and to facilitate for creating a new die geometry by reusing or revising the die features from library. The die features can be easily modified in order to use the information from features for supporting the process planning of die manufacturing. This research die features can be divided into two categories; die part geometry and die feature library. Die part features compose and describe the feeder plate, die plate, backer plate, mandrel, die cap, back plate, bolster and sub-bolster. These part features can be reused for new die design by adding the required feature from die feature library. Die parts feature library are illustrated in Figure 18.

These systems geometry are created by computer-aided design such as SolidWorks®. It has feature library to facilitate and reuse the standard features in order to modify geometry. Thus feature library in CAD software can be applied to support die design based on feature base design. Feature library of die geometry can be used to modify feed feature, die orifice, backer opening, porthole, welding chamber, and so on. Moreover, each feature has entities to be used in machining process or computer aided manufacturing.

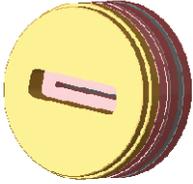

Figure 17: Feature class library

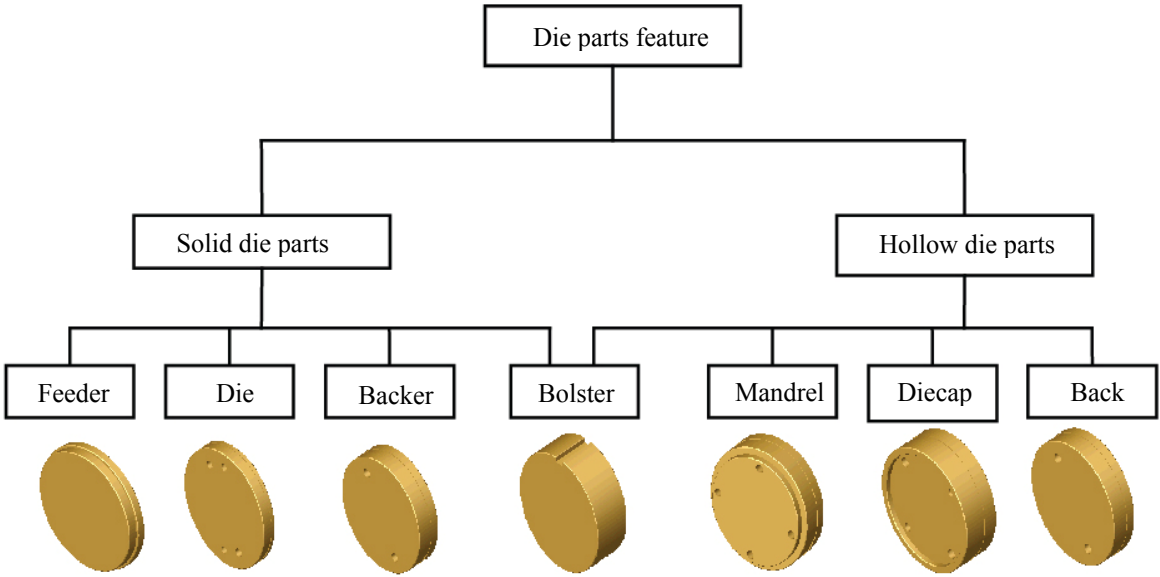

Figure 18: Die part feature library

### 4.3.1 Feeder feature

Feeder has function to control the metal flow through die orifice. For solid die, feed is in the feeder plate or die plate whereas for hollow die, feed is the internal part of the mandrel and is called porthole. Feeder shape affects the flow pattern, the extrusion pressure, the metal flow velocity, and the consequence the bearing length. Bearing length is adapted, based on the local wall thickness and the velocity of the metal flow just at the entry through the die orifice. So the bearing length relates directly with feeder shape. For instance in Figure 18 shows typical feed feature library and simple holes feature. Die part feature as feeder plate can be subtracted with feeder feature as dog bone shape as shown in Figure 19. A user can modify feature dimensions to attain a new feed geometry for each design case based on profile geometry.

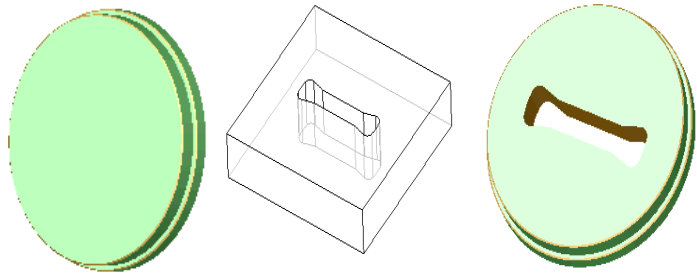

Figure 19: Typical feed and simple holes feature

### 4.3.2 Welding chamber feature

Metal flows from each porthole of mandrel in hollow die and joins again before to be pushed through die orifice in a welding chamber. Welding chamber geometry is designed depending on porthole design and welding line allocation on the section. In practice, welding line should be avoided allocation on the exposed surface. A die designer can select the welding chamber feature in die feature library. Welding chamber feature can be modified in order to obtain the required shape and dimensions. Moreover, the welding chamber depth is given according to the wall thickness of the section and is determined from the total die depth. Figure 20 illustrates typical welding chamber features for using it to create the welding chamber in die cap of hollow die geometry.

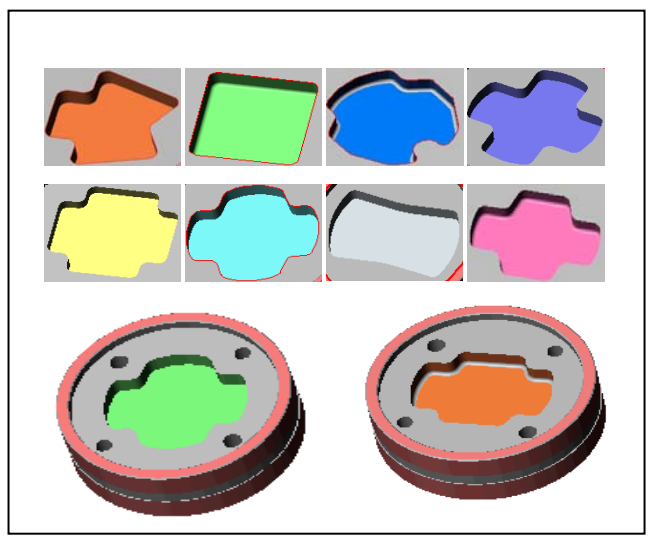

Figure 20: Typical welding chamber features

## 5 PROCESS PLANNING OF ALUMINUM EXTRUSION DIES

### 5.1 Process planning reviews

Process planning is defined as the planning and development of detailed instructions for the conversion of a raw material into a finished part based on some feasible engineering design [7]. Process planning encompasses the activities and functions to prepare a detailed set of plans. It includes the sequence of operations, selection raw material, machines, tools, cutting tools, cutting parameters, etc. In general, process planning is based on the skill and experience of a process planner who can provide the knowledge of process planning. However, the human memory may be lost and it is not easy to collect enormous data. Knowledge based system has been determined to manage the implicit and explicit knowledge from any sources in a computer system. In recent years, artificial intelligence has been discussed and used in order to develop in a computer system for supporting process planning tasks. In general process planning can be distinguished into three methods:

- The VPP (Variant Process Planning) method
- The generative method
- The hybrid method

The variant process planning involves to retrieve an existing plan for similar part and to modify the previous plan for the new part. Variant method is based on a Group Technology (GT), coding and classification approach to identify a larger number of part attributes or parameters. The structure of VPP begins by coding the part by a user and then searches the similar part from part database. Each part of die set has a standard machine routing and retrieve the standard operations from operations sequence file for reusing and modifying in order to apply with a new part. Finally, the system gives process plan. However, the machining routing of standard parts can be generated from the knowledge of a die process planner.

The generative method is based on useful knowledge base, production rules, or artificial intelligence to generate process plans. This method approaches deals with generation of new process plans by means of decision logics and process knowledge. Data from CAD is extracted to recognize the feature shapes and retrieve the dimensions for generating the process plans with an inference engine. The inference engine decides the selection process and machine by deriving from the knowledge base. In fact, knowledge base for process planning comes from the acquisition of the knowledge from experts, from the machine specifications and capabilities, and so on. In addition, knowledge base provides the operations sequence, machines, cutting tools, cutting parameters, machining times, etc via the inference engine. To realize a generative process-planning module, it is necessary to have a knowledge base which includes three main components: the part description, the knowledge base and database, and the decisional logics and calculus algorithm.

The hybrid method approach attempts to exploit knowledge in existing plans while generating a process plan for a new design [8]. Computer aided process planning can eliminate many of the decisions required during planning. It has the following advantages: reduces the demand on the skilled planner, reduces the planning time, reduces process planning and manufacturing costs, creates consistent plans, produces accurate plans, and increases productivity [9]. The knowledge-based expert systems for process planning are reviewed in [10]. By example, it is used to be application for in research [11] and [12]. In addition, Stryczek reviewed the application of computational intelligence in computer aided process planning [13].

### 5.2 The knowledge of process planning

The knowledge base is the central component of an expert system. It contains generic knowledge for solving specific domain-related problems. In general, we can classify the knowledge base into three main categories. There are procedural knowledge, declarative knowledge, and control knowledge. Procedural and declarative knowledge are stored in the knowledge base. The control knowledge is used to construct the inference mechanism. Procedural knowledge can be sometimes also called production rules which represent the relation structure and problem oriented hierarchies of the knowledge stored or the production rules. The declarative knowledge may be called the knowledge details or problem facts, represents the factual part of the knowledge or the specific features or the problems. The knowledge may be a group of data or a symbolic structure. Additionally, the declarative knowledge can be stored in a database. The control knowledge is the knowledge about a variety of processes, strategies, and structures used to coordinate the entire problem–solving process, is not stored in the knowledge base. The knowledge base for process planning is used to store the production rules, an inference engine, a database, and so on. It can be used in order to perform the process planning for aluminum extrusion die. The knowledge base of process planning and cost evaluation contains rules and techniques for knowledge representation. It includes cutting tools, component features and machining process [14]. The proposed knowledge-based system utilizes (rules, tables, equations, etc.) for: component specification, tool material selection, machining process, cutting tools selection, and cutting conditions. Martin and D'Acunto [15] presented a procedure for the design of a production system, based on part modeling and formalization of technological knowledge by using production features, and finally they can calculate production costs by determining from the multi-criteria analysis. Grabowik and Knosala [16] presented a method of representation of the knowledge about the body construction and technology in an expert system that aids the process of designing the machining technology of bodies. Expert system is focused on the improvement of the proposed method of object representation on the technological knowledge and body construction in order to increase the level of the generated technological documentation by increasing the number of the problems. For this research we employed frame-based system and rule-based in order to manage the knowledge base of die process planning and cost estimating.

**5.3 Die machining process**

Machining processes for making a part of die consist of various processes depending on geometry of each part as following: turning, drilling, milling, EDM (wire cutting, drilling, sparking), grinding, and assembling. Moreover, heat-treatment is the importance method to improve mechanical properties of die material for supporting hot working and high pressing.

**5.3.1 Turning process**

Turning process is the early stage to form raw material cylindrical shape to obtain die disc before machining with another processes. Facing, turning, and chamfering are the operations for machining each die part. Turning operation of a die is shown in Figure 21. A measurement of how fast material is removed from a work piece can be calculated by multiplying the cross section area of the chip by the linear travel speed of the tool along the length of the work piece. Material removal rate in turning can be calculated in the form:

$$\mathbf{MRR} = \pi \times \mathbf{D} \times \mathbf{d} \times \mathbf{f} \times \mathbf{N} \tag{1}$$

Where:
    $D$   is outer (or average) work piece diameter
    $d$   is depth of cut
    $f$   is feed rate (ipr)
    $N$   is spindle speed (rpm)

The spindle speed is given by:

$$N = \frac{12 \times v}{\pi \times D} \tag{2}$$

Material removal rate becomes:

$$\mathbf{MRR} = 12 \times \mathbf{d} \times \mathbf{f} \times \mathbf{V} \tag{3}$$

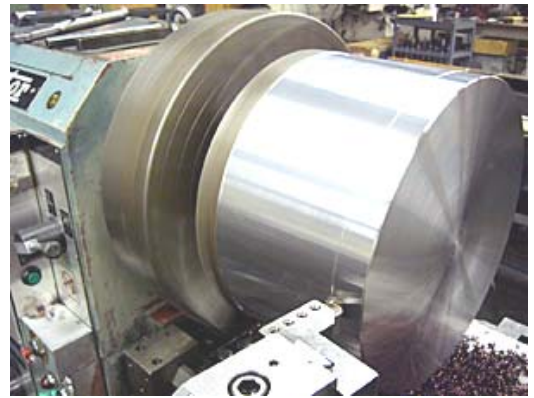

Figure 21: Turning operation

Where:
    $V$   is cutting speed (fpm)

In straight turning, chip width is $\left(\dfrac{D_0 - D_f}{2}\right)$

Where:
    $D_o$ and $D_f$ are outer and inner work piece diameters, respectively

Cross section area is $\dfrac{\pi(D_0^2 - D_f^2)}{4}$ hence the material removal rate is

$$MRR = \frac{\pi(D_0^2 - D_f^2)}{4} \times f \times N \tag{4}$$

**5.3.2 Milling process**

Milling is widely applied in machining process due to it includes a number of highly versatile machining operations capable of producing a variety of configurations. The basic types of milling such as slab milling, face milling, end milling, and so on. Milling process can be machined feed, porthole, weld chamber, etc. for die part machining as shown in Figure 22.

Material removal rate in milling process can be calculated by determination from milling parameters as the cutting speed, *V (m/min)*, is given by:

$$V = \pi \times D \times V \quad (5)$$

Where:
    *D* is the cutter diameter (mm)
    *N* is the rotational speed of the cutter (rpm)
Feed per tooth, mm/tooth can be calculated by:

$$f = \frac{v}{N \times n} \quad (6)$$

Where:
    *v* is linear speed of the work piece or feed rate, mm/min
    *n* is number teeth on cutter
Then material removal rate is given by the expression

$$MRR = w \times d \times v \quad (7)$$

or

$$MRR = w \times d \times (f \times N \times n) \quad (8)$$

where:
    *w* is width of cut (mm)
    *d* is depth of cut (mm)

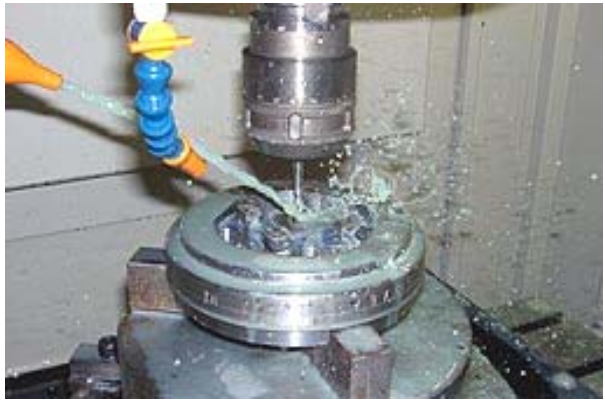

Figure 22: Milling operation

### 5.3.2 Grinding process
Grinding process is a chip-removal process that uses an individual abrasive grain as cutting tool. The grinding process is used to flatten the die surface in order to assemble and to correct die distortion. Die face must be flatness to avoid unbalance of the metal flow. Material removal rate in grinding machining is given by:

$$MRR = d \times w \times v \quad (9)$$

Where:
    *d* is depth of cut (mm)
    *w* is width of cut or grinding wheel thickness (mm)
    *v* is linear speed of the work piece or feed rate, mm/min

### 5.3.3 Wire EDM machining process
Electrical discharge wire cutting is used to cut contour as thickness plates, punches, dies, and small or complex holes. It will be cut die orifice or deep hole in die manufacturing process. The wire is a cutting tool that is usually made of brass, copper, or tungsten; zinc or brass coated and multi-coated wires are also used. The wire diameter is typically about 0.30 mm for roughing cuts and 0.2 mm for finishing cuts. The cutting speed is generally given in terms of the cross-sectional area cut per unit time. Typical examples are: 18,000 mm$^2$/hr for 50 mm thick D2 tool steel, and 45,000 mm$^2$/hr for 150 mm thick aluminum. These removal rates indicate a linear cutting speed of 18,000/50 is 360 mm/hr or 6 mm/min, and 45,000/150 is 300 mm/hr or 5 mm/min, respectively. Typical EDM wire cutting process is illustrated in Figure 23.

### 5.3.3 Electro discharge machining process
Electro-discharge or spark-erosion machining, is based on the erosion of metals by spark discharges. The basic EDM system consists of a shaped tool (electrode) and work piece, the metal surface is removed from a transient spark discharges through the dielectric fluid. Electrodes for EDM are usually made of graphite, brass, copper, and copper-tungsten is also used. EDM sparking is generally used to make bearing length of an extrusion die. Metal removal rates usually range from 2 to 400 (mm$^3$/min).

Moreover, EDM can be used to make the core shape of mandrel as shown in Figure 24.

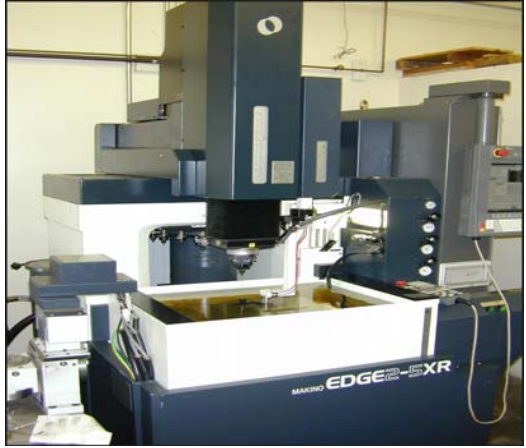

Figure 23: EDM wire cutting process

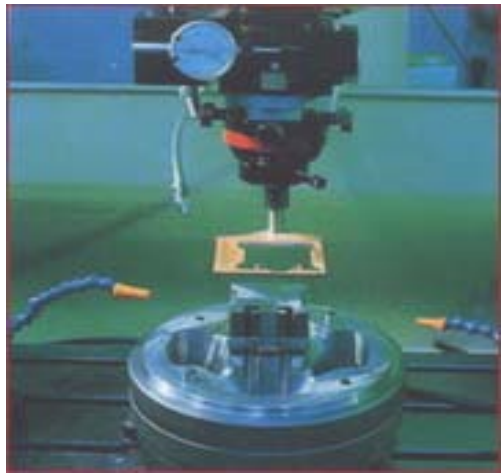

Figure 24: EDM sparking process

### 5.3.4 Heat treatment
In addition, the important process in die manufacturing is heat treatment. This process has played a role to increase die strength property.

From machining process as described above for fabrication an extrusion die, we proposed machining process classes as turning process class, drilling process class, milling process class, and so on are derived from machining process class. These classes have methods to calculate machining parameters and metal removal rate in order to calculate machining time of each process.

## 6 THE KNOWKEDGE BASE OF DIE DESIGN AND PROCESS PLANNING

### 6.1 The structure of knowledge based system for die manufacturing
For our application, we employed frame-based system and rule-based in order to manage the knowledge base of die process planning and cost estimating. Figure 25 presents the flow chart of the knowledge base process planning and cost estimating of a die. The knowledge is managed in frame-based system.

To approach process planning of a die, we need form features and material type of die part in order to select the feasible machining processes. Die features are retrieved by two ways: from die part geometry or die features database.

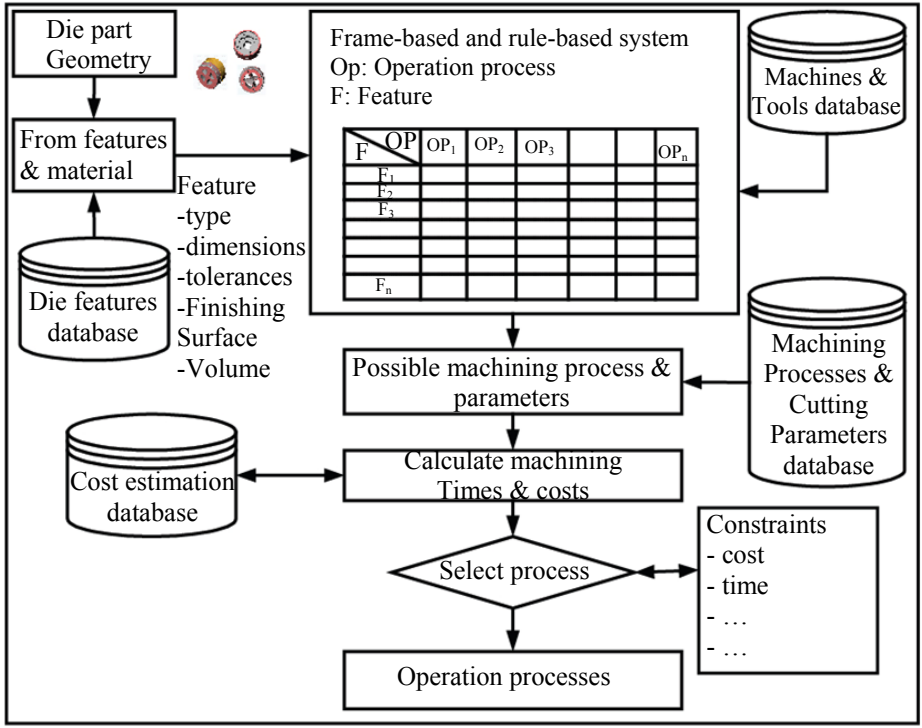

Figure 25: Process planning and cost estimating flow chart of an aluminum extrusion die

On one hand, a user has to define all features of their parts into a database of a new die part case. On the other hand, we can reuse die features by reusing previous die part features from database. Die feature includes feature class type, dimensions, tolerances, finishing surface, volume, etc. These features are then used to select machining process by frame-based, rule-based system and decision table. Each feature has different operation processes to be machining. The system gives the possible machining processes by frame-based and select the suitable those processes based on rule-based system and decision table. The selected process composes of operation method, machine type, machine tools, and cutting parameters. Machines and tools data can also be retrieved from machine and tools database. Machining process and cutting parameters database provide die operation method in the process and cutting conditions in order to calculate cutting time. Each machining process will be evaluated cutting time by formulas and machining time is then used to calculate machining cost. Finally, the system selects the process according to constraints as machining time and cost.

This knowledge base system for process planning of the extrusion dies is used to train the data sets of input and output in neural network architecture. The next section describes the structure of die process planning with artificial neural network technique.

### 6.2 The structure of die design and process planning system with artificial neural network

The structure of die design and process planning is illustrated in Figure 26. This structure consists of two main parts, including die design and process planning. The first part is design die features based on the neural networks as defined in 7.1. The process starts by reading product data from database or from customer's product. The product data of aluminum extrusion process is the characteristics of profile. It contains profile shape, section area, dimensions, tongue ratio. This data is input data layer of the neural network in order to search the previous die design cases from die design case library. The output layer is the similar case which is retrieved from library to be reused the features of die design geometry. The machining processes for making the die features can be given by the knowledge base of die machining process planning.

The frame based and ruled based systems are managed and derived the machining processes by determining the machined features of die geometry. Tools database can support the system to select the machine tools for each machining operation. Process parameters are given from process database. These machining parameters are cutting speed, cut of depth, feed rate, and so on. They can be defined with the knowledge base of the suggestion from tool's manufacturer and shop floor data of die maker. Moreover, die machining process sequencing is ordered according to die making routes standard. Each machining process route depends on the type of die part. Eventually, machining process plan is generated by the proposed structure of die manufacturing based on neural network technique.

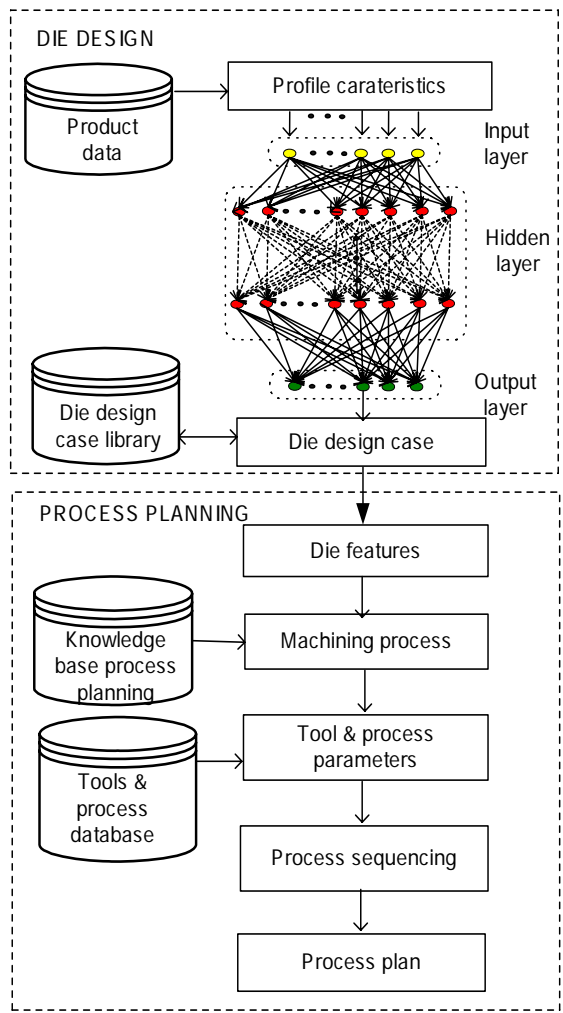

Figure 26: The structure of die manufacturing with artificial neural network

# 7 FEATURE BASED NEURAL NETWORK FOR ALUMINUM EXTRUSION DIE DESIGN AND MANUFACTUTING

## 7.1 Artificial neural network in CAPP

In last decade, artificial neural network has been widely used in engineering application domain, especially for design and manufacturing. Artificial neural network is a mathematical model for parallel computing mechanisms as same as biological brain. They consist of nodes, linked by weighted connections. Neural networks are constructed by hierarchical layers, which are input, hidden, and output layer respectively. Neural networks learn relationships between input and output by iteratively changing interconnecting weight values until the outputs over the problem domain represent the desired relationship. Furthermore, neural networks perform a variety of functions such as pattern matching, trend analysis, image recognition, and so on. CAPP (Computer Aided Process Planning) is the important task to couple CAD and CAM by interpreting feature model to machining process. Knapp et al [17] presented the ability of neural network in the process selection and within feature process sequencing. In this work, two co-operating neural networks were utilized: the first one, a three layer back propagation neural network, takes in as input the attributes of a feature and proposes a set of machining alternatives; another fixed weight neural network selects exactly one of the alternatives. Parameters of the features are modified by the results of the operation until the final state of the feature has been reached. Yahia & al [18] proposed a feed forward neural network based intelligent system for computer aided process planning methodology. This methodology suggests the sequence of manufacturing operations to be used, based on the attributes of a feature of the component. By integrating this methodology with computer aided design (CAD), process planning can be generated, and tested, which helps in realizing concurrent engineering. Yue and al [19] presented a state of the art review of research in computer integrated manufacturing using neural network techniques. Neural network-based methods can eliminate some drawbacks of the conventional approaches, and therefore have attracted research attention particularly in recent years. The four main issues related to the neural network-based techniques, namely the topology of the neural network, input representation, the training method and the output format are discussed with the current systems. The outcomes of this research using neural network techniques are studied, and the limitations and future work are outlined. Praszkiewicz [20] purposed of this article is to present the application of neural network for time per unit determination in small lot production in machining. A set of features considered as input vector and time consumption in manufacturing process was presented and treated as output of the neural net. A neural network was used as a machining model. Sensitivity analysis was made and proper topology of neural network was determined.

## 7.2 Structure of artificial neural networks for die design and process planning

The proposed structure of neural network for die design and process planning is shown in Figure 27. The input parameters consist of the characteristics of aluminum profile including type of profile, profile shape, dimensions, cross-section area, extrusion ratio, CCD (Circumscribing Circle Diameter), tongue ratio etc. The output layer contains the type of die, the number die orifice, extrusion ratio, die stack set, and die machining process routes.

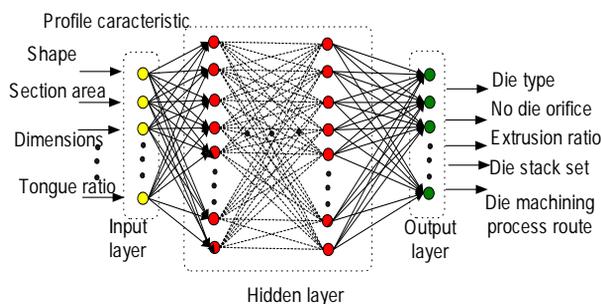

Figure 27: The structure of neural networks for die design and process planning

The mathematical model of the biological neuron, there are three basic components as presented in Figure 28.

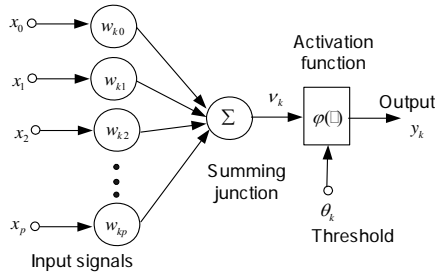

Figure 28: A perceptron neuron model

First, the synapses of the neuron are modeled as weights. The value of weight can present the strength of the connection between an input and a neuron. Negative weight values reflect inhibitory connections, while positive values designate excitatory connections. Second component is the actual activity within the neuron cell. This activity is referred to as linear combination. Finally, an activation function controls the amplitude of the output of the neuron. An acceptable range of output is usually between 0 and 1, or -1 and 1.

Each neuron calculates two functions. The first is propagation function as shown in equation 10,

$$v_k = \sum w_{kj} x_j \qquad (10)$$

Where $w_{kj}$ is the weight of the connection between neuron $k$ and $j$, $y_k$ is the output from neuron $k$. The second is an activation function. The output of a neuron in a neural network is between certain values (usually 0 and 1, or -1 and 1). In general, there are three types of activation functions, denoted by $\varphi(\bullet)$ as illustrated in equation 10. Firstly, there is the treshold function which takes on a value of 0 if the summed input is less than a certain threshold value ($v$), and the value 1 if the summed input is greater than or equal to the threshold value.

$$\varphi(v) = \begin{cases} 1 & \text{if } v \geq \theta_k \\ 0 & \text{if } v < \theta_k \end{cases} \qquad (11)$$

Training of the networks will be discussed in the next section.

## 8  DATASET FOR NETWORK TRAINING

The key issues of the developed neural network based methodology for aluminum extrusion die design and process planning of die manufacturing will be discussed in the following sections:

### 8.1  Formulate the knowledge based system of die design and process planning

A set of rules has been generated to define the feature of die features and machining processes for each feature. These rules have been used to be the set of input and output layer of the neural network structure. These rules are captured from the successful die design and machining cases which are stored in die design case library. The rules of thumbs may be gathered with the knowledge from die design and die making experts and other source. The rules are formed in IF-THEN.

### 8.2  Design topology of the neural network model

The neural network has been trained by using the standard propagation algorithm. This work uses supervised learning, which is one of three categories of the training method. Supervised learning may be called associative learning, is trained by providing with input and matching output patterns. These input-output pairs can be contributed by an external teacher, or by the system which contains the neural networks (self-supervised). The learning process or knowledge acquisition takes place by representing the network with a set of training examples and the neural network via the learning algorithm implicitly rules. The topology of the proposed neural network model applies feed forward architecture. Each variable is the input value at a node of the input layer. The input layer of neuronal node is designed in such a way that one node is allocated for the feature type, and one node is allocated to each of the above sets of feature attributes. Also the values of all the input layer neurons are normalized to lie between 0 and 1. The number of nodes in the input layer is equal to one plus the number of all the possible different ranges of feature attributes, encountered in the antecedent part of the rules. The types of profile are represented by 3 nodes to denote solid, semi-hollow and hollow type respectively. Profile shape is also given 20 principle shapes which are addressed with 20 nodes. In addition, general profile thicknesses are designed with 20 nodes, and each node has the difference value ranges in 0.5 mm. Moreover, dimensions (maximum width and height), CCD, cross section area, press machine capacity, extrusion ratio, perimeter, external perimeter and tongue ratio define the numbers of the nodes in input layer which are 15, 15, 15, 15, 3, 30, 15, 17 and 12 respectively in our case. Consequently, the total number of nodes in input layer is 170. For example, the typical heat sink profile is shown in Figure 29. The characteristics of profile consist of: solid profile with rectangular shape, general wall thickness 2.3 mm, width 24 mm, height 15.3 mm, CCD 28.5 mm, section area 1.7 cm$^2$, perimeter 20.32 mm, without external perimeter, and tongue ratio 4.0.

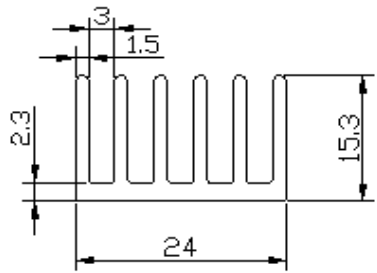

Figure 29: Typical aluminum profile

We can transform the characteristics of the typical profile to input layer of the network in vector format as illustrated in Table 2.

Table 2: The typical vector of input layer

| Column | 1 | 2 | 3 | 4 | 5 | 6 | 7 | 8 | .. | 170 |
|---|---|---|---|---|---|---|---|---|---|---|
| Value | 1 | 0 | 0 | 0 | 0 | 0 | 0 | 0 | | 0 |

In the above vector, the column numbers [1-3] addresses the type of profile, [4-23] stand the profile shapes, and [24-33] stand for the sets corresponding to the different ranges of general wall profile thickness. Column numbers [34-48], [49-63], and [64-78] are addressed according to the different ranges of maximum width, height dimensions, and CCD respectively. Column numbers [79-93] are the ranges of cross section area. Column numbers [94-96] address the press machine capacity as 660, 880, and 1800 tones respectively. Column numbers [97-126] stand for the sets corresponding to extrusion ratio. Column numbers [127-141] and [142-158] are presented the ranges of perimeter and external perimeter. The last column numbers [159-170] stand the ranges of tongue ratio.

The output decision variables for the die design and process planning comprise of the various feature of die geometry and die machining process plan. In addition, process sequencing and machining operations are given in the output layer. Die features and machining process are derived from the knowledge based system of die design and process planning. The first part of output network is a kind of dies, including 3 nodes as solid, semi-hollow, and hollow die as same as the third 3 nodes input. The next two groups are the nodes of the number die orifice 15 nodes and the nodes of extrusion ratio 15 nodes respectively. Die set thicknesses are also designed corresponding to die stack dimensions of each press machine. Die thickness consists of feeder plate, die plate, backer plate thickness for solid or semi-hollow die, and mandrel, die cap thickness for hollow die. The number of die thickness group is 50 nodes. Die features are defined following the main types of die components, including feeder, die, back, mandrel and die cap part features. These features consist of four main feature categories, are hole, edge, groove, and pocket. Hole features have blind, through, tap hole, counter sink, counter bore, and deep hole. Edge features include edge chamfer and edge fillet. Groove features are v groove, round groove, and rectangular groove. The last group is pocket which comprises open pocket plane, open pocket circular, open pocket sculptured, closed pocket plane, closed pocket circular, and closed pocket sculptured. The number node of die features can be classified into 5 nodes following as die part components. The first node is feeder features which have open pocket circular, open pocket plane, edge chamfer, tap, and close pocket plane. The second node addresses die features, including open pocket circular, open pocket plane, edge chamfer, through, deep hole, and close pocket plane. The third node stands back features which are open pocket circular, open pocket plane, counter bore, and close pocket plane. The fourth node presents mandrel features, comprising open pocket circular, open pocket plane, edge chamfer, tap, open pocket sculptured, and close pocket sculptured. The last node of die feature group nodes is die cap feature, including open pocket plane, open pocket circular, edge chamfer, counter bore, close pocket plane and deep hole.

The last part of output layer is die parts machining processes (turning, drilling, milling, heat treatment, grinding, EDM (sparking, wire cutting, drilling) process). Turning process includes rough turning, semi-finish, finish turning, round chamfering, and round grooving operation. In addition, facing process is discussed to remove the material of open pocket circular in order to control die part thickness. Rough facing, semi-finish facing, and finish facing are operations of facing process. Drilling is the simple process to make a hole in die geometry. There are seven operations: centering, drilling, boring, reaming, tapping, counter boring, and countersinking. The important machining process to be cut the open pocket plane, and open pocket sculptured at mandrel, feeder plate, and back orifice of back plate is milling process. This process comprises rough, semi-finish, and finish milling operation. These machining processes and operations are also grouped in five nodes according to die machining process planning routes which have feeder, die, back, mandrel, and die cap machining process routes respectively.

In addition, for aluminum extrusion manufacturing, heat treatment process is applied to enhance die strength property. After heat treatment, die may be distorted then it must be machined with grinding process to flatten die face. EDM sparking process is used to make bearing lengths only die plate and die cap. The last process is EDM wire cutting which includes rough and finish wire cutting. Die part of solid die and die cap of hollow die are used wire cutting process to cut die orifice. So the number of nodes in the output layer is 93. The output layer vector is shown in Table 3.

Table 3: The typical vector of output layer

| Column | 1 | 2 | 3 | 4 | 5 | 6 | 7 | 8 | .. | 93 |
|--------|---|---|---|---|---|---|---|---|----|----|
| Value  | 1 | 0 | 0 | 1 | 1 | 1 | 1 | 1 |    | 0  |

Tolerance and surface finish are not seriously determined in aluminum extrusion die manufacturing due to the machining process selection are not based on the tolerance and surface finish value. The process routes of die machining operation are consistency, it can be set the standard process.

In the above vector of output layer, the column numbers [1-3] stand the type of die. The column numbers [4-18] address the number of die orifice and extrusion ratio is presented in column numbers [19-33]. The column numbers [34-43], [44-53], [54-63], [64-73], and [74-83] stand the part thickness of feeder, die, back, mandrel and die cap respectively. The next group column numbers [84-88] are the feature group of die part features. Finally, the last 5 nodes as shown in column number [89-93] stand the machining process routes in order to make die parts. They are feeder, die, backer, mandrel, and die cap process planning routes respectively.

### 8.3 Training the neural network

*S. Butdee et al.*

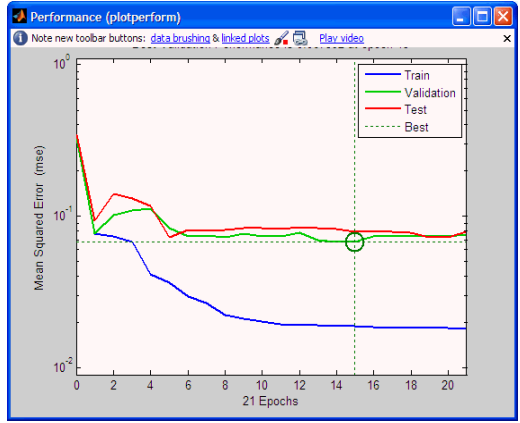

Figure 30: Learning curve

The standard back-propagation algorithm is used as the learning mechanism for the neural network. The training data set has been prepared by translate die design cases data from cases library to input and output layer format. Neural network tool box of MATLAB 2008 is employed to simulate the neural network operation. There are two alternative training modes possible, depending on the particular way of presenting the training patterns to the neural network and depending on when the network weights are updated, either after presentation of each training pattern or after presentation of the entire set of examples. For this paper, we found that the number of hidden layers is 1. The mode of training is pattern. The number of hidden layer is 5. Learning rate is o.1 and momentum rate is 0.7 respectively. The numbers of nodes in input and output layer are 170 and 93. The learning curve is shown in Figure 30.

## 9  CASE STUDY

This case study is tested with the aluminum extrusion die design and process planning to make a die from the example case in an aluminum extrusion industry. The sample product is selected to test the developed system. The sample product is created by CAD system in order to extract the characteristics of this profile for use in die design and process planning a die. The product is shown in Figure 31.

The characteristics of this profile are including hollow profile with rectangular shape, cross section area is 3.4 $cm^2$, profile width is 50 mm and height is 14.7 mm, perimeter is 30.37 cm, and external perimeter is 19.24 cm, and tongue ratio is 1.4 respectively. This product data is transformed to input layer format in order to find out the die features and process planning. The result of this case is illustrated in Table 4 that has been translated from the codes of the output layer to features of a die and process planning.

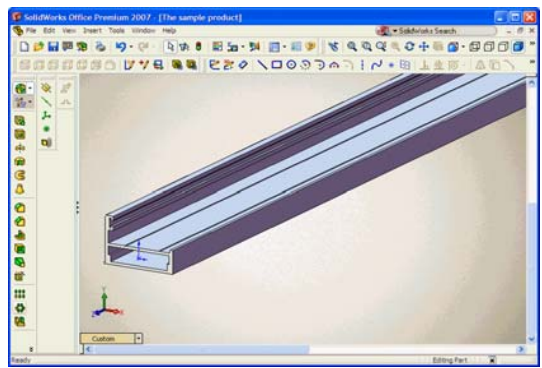

Figure 31: The sample product in CAD system

The type of die is hollow die with one cavity. Extrusion ratio is 40. Die components are mandrel and die cap. There fore, the machining process plans are mandrel, and die cap process planning route as shown in Table 4.

Table 4: The features and machining process plan of the die design case study

| Die part | Features | Machining process and operations |
|---|---|---|
| Mandrel | Open pocket circular | Rough turning<br>Semi-finish turning<br>Finish turning<br>Rough facing<br>Semi-finish facing<br>Finish facing |
| | Edge chamfer | Round chamfering |
| | Blind hole | Centering<br>Drilling |
| | Tap | Tapping |
| | Closed pocket sculptured | Rough milling<br>Semi-finish milling<br>Finish milling |
| Die cap | Open pocket circular | Rough turning<br>Semi-finish turning<br>Finish turning<br>Rough facing<br>Semi-finish facing<br>Finish facing |
| | Blind hole | Centering<br>Drilling |
| | Counter bore | Centering<br>Drilling<br>Counter boring |
| | Closed pocket sculptured | EDM Sparking |
| | Close pocket plane | Rough milling<br>Semi-finish milling<br>Finish milling |
| | Deep hole | Rough wire cutting<br>Finish wire cutting |

## 10  SUMMARY

This paper presents the result of application feature based method with artificial neural network in order to search the die design and process planning case from aluminum extrusion die manufacturing library. The outputs are die design features and possible die machining process plan. The die machining process plan consists of machining operations and sequences of each process. The detailed description of the neural network based methodology includes formulating the knowledge base system for die design and process planning, designing topology of the neural network model, and training the neural network have been given. The knowledge based system of die design and manufacturing are acquired from the die design and manufacturer experts with other sources such as textbooks, researches, and so on. This knowledge is used to train in the neural network structure. The potential for application of the developed feature based neural network model has been presented with the help to design and plan the machining process of an aluminum extrusion die. Die design case study is successfully tested with the developed system. The number of die manufacturing cases from the aluminum extrusion industry is more than 150 cases. The lead time of die design and process planning are decreased, and increase the efficiency of die design and manufacturing to make a die to support extrusion process. It can enhance the competitive in order to produce a good aluminum extruded profile.

In addition, the fundamental of aluminum extrusion process, die design and process planning are described to understand the principal knowledge of aluminum extrusion process.

## 11 ACKNOWLEDGMENTS

We would like to thank MTAlumet Co., Ltd. Thailand for supporting the information and the resources to be used for this research. In addition, the authors thank Thai research fund and French government for funding our project.